\documentclass[journal]{IEEEtran}
\IEEEoverridecommandlockouts

\usepackage{cite}
\usepackage{amsmath,amssymb,amsfonts}
\usepackage{algorithm}
\usepackage{algorithmic}
\usepackage{graphicx}
\usepackage{textcomp}
\usepackage[table,xcdraw]{xcolor}
\usepackage{multicol}
\usepackage{cuted}
\usepackage{extarrows}
\usepackage{booktabs}
\usepackage{caption}
\usepackage{multirow}
\usepackage{lipsum}
\usepackage{hyperref}
\usepackage{nicematrix}
\usepackage{amsthm}
\theoremstyle{definition}

\usepackage{subcaption}
\usepackage{tikz}
\usetikzlibrary{shapes.geometric, arrows.meta, positioning}

\definecolor{headercolor}{rgb}{0.369, 0.788, 0.478}
\definecolor{positivecolor}{rgb}{0.494, 0.753, 0.933}
\definecolor{negativecolor}{rgb}{0.988, 0.553, 0.384}

\def\BibTeX{{\rm B\kern-.05em{\sc i\kern-.025em b}\kern-.08em
    T\kern-.1667em\lower.7ex\hbox{E}\kern-.125emX}}
\begin{document}

\title{Predicting Market Trends with Enhanced Technical Indicator Integration and Classification Models}

\author{
Abdelatif Hafid, Abderazzak Mouiha, Linglong Kong, Mohamed Rahouti, Maad Ebrahim,\\
Mohamed Adel Serhani, and Mohammed Aledhari%

\thanks{Manuscript received xx, xx; revised xx, xx.}

\thanks{Abdelatif Hafid and Abderazzak Mouiha are with the School of Digital Engineering and Artificial Intelligence, Euro-Mediterranean University of Fes (UEMF), Fes 30090, Morocco (e-mail: a.hafid@ueuromed.org).}
\thanks{Linglong Kong is with the Department of Mathematical and Statistical Sciences and the Alberta Machine Intelligence Institute, University of Alberta, Edmonton, AB T6G 2G1, Canada.}
\thanks{Mohamed Rahouti is with the Department of Computer and Information Science, Fordham University, New York, NY 10023, USA.}
\thanks{Maad Ebrahim is with Department of Computer Science and Operations Research, Université de Montréal, Montreal, Canada.}
\thanks{Mohamed Adel Serhani is with the Department of Information Systems, University of Sharjah, UAE.}
\thanks{Mohammed Aledhari is with the Department of Information Science, University of North Texas, Texas, USA.}
}

\maketitle

\begin{abstract}
Thanks to the high potential for profit, trading has become increasingly attractive to investors as the cryptocurrency and stock markets rapidly expand. However, because financial markets are intricate and dynamic, accurately predicting prices remains a significant challenge. The volatile nature of the cryptocurrency market makes it even harder for traders and investors to make decisions. This study presents a classification-based machine learning model to forecast the direction of the cryptocurrency market, i.e., whether prices will increase or decrease. The model is trained using historical data and important technical indicators such as the Moving Average Convergence Divergence, the Relative Strength Index, and the Bollinger Bands. We illustrate our approach with an empirical study of the closing price of Bitcoin. Several simulations, including a confusion matrix and Receiver Operating Characteristic curve, are used to assess the model's performance, and the results show a buy/sell signal accuracy of over 92\%. These findings demonstrate how machine learning models can assist investors and traders of cryptocurrencies in making wise/informed decisions in a very volatile market.
\end{abstract}


\section{Introduction}  \label{sec:introduction} 

The cryptocurrency market has experienced explosive growth in recent years, transforming stock trading into an attractive financial option for investors because of its high profitability and accessible entry. However, investing in stocks has inherent risks, highlighting the importance of developing a well-defined investment plan/strategy. Traditionally, investors relied on empirical approaches (e.g., technical analysis) guided by financial expertise. However, with the widespread integration of financial technology (FinTech), statistical inference models have emerged, leveraging machine learning techniques to forecast/predicting stock price movements. This change in approach has shown notable success on various stock markets, including the S\&P 500, NASDAQ \cite{hsu2021fingat}, and cryptocurrency market \cite{liu2021forecasting, hafid2023bitcoin}.

This research focuses on the dynamic and rapidly evolving cryptocurrency market, particularly on Bitcoin price prediction \cite{nakamoto2008bitcoin}. The choice of this focus is motivated by several factors. Firstly, the cryptocurrency market represents a frontier in financial innovation, challenging traditional currency and value storage notions. Secondly, the underlying blockchain technology that powers cryptocurrencies has attracted significant attention from the banking and financial sectors due to its potential to revolutionize transaction processing and data management \cite{hafid2020scaling}.

Despite the numerous advantages of the cryptocurrency market, including abundant market data and continuous trading, it presents unique challenges that warrant in-depth investigation. These challenges include:

\begin{enumerate}
    \item High price volatility: Cryptocurrency prices are subject to rapid and significant fluctuations, making accurate prediction particularly challenging.
    \item Relatively small market capitalization: Compared to traditional financial markets, the cryptocurrency market's smaller size can increase susceptibility to market manipulation and external influences.
    \item Data analysis complexity: While data accessibility is universal in cryptocurrency trading, success hinges on effectively analyzing and selecting relevant information from vast datasets.
\end{enumerate}

The primary challenge in cryptocurrency trading lies in distinguishing between successful/profitable and unsuccessful/unprofitable trades in this complex environment. Consequently, developing sophisticated machine learning models capable of extracting meaningful insights from available data is paramount. Moreover, the increased price volatility inherent to cryptocurrencies adds a layer of complexity to price forecasting efforts.

Machine learning models, such as Long Short-Term Memory (LSTM) networks and Random Forests (RF) \cite{hafid2023bitcoin}, have emerged as powerful tools in addressing these challenges, particularly in predicting cryptocurrency prices. These models leverage historical data and identify complex patterns to make informed predictions, thereby contributing to more effective decision-making in the dynamic landscape of cryptocurrency trading.

\begin{figure}[!ht]
    \centering
    \begin{tikzpicture}[node distance=1cm and 1cm]
    \node (start) [draw, rectangle, rounded corners, minimum width=2.5cm, minimum height=1cm, text centered, text width=2.5cm, fill=white!10] {1. Research Problem};
    
    \node (data) [draw, rectangle, minimum width=2.5cm, minimum height=1cm, text centered, text width=2.5cm, below=of start, fill=white!10] {2. Data Collection};
    
    \node (preprocess) [draw, rectangle, minimum width=2.5cm, minimum height=1cm, text centered, text width=2.5cm, below=of data, fill=white!10] {3. Feature Engineering};
    
    \node (features) [draw, rectangle, minimum width=2.5cm, minimum height=1cm, text centered, text width=2.5cm, below=of preprocess, fill=white!10] {4. Data Preprocessing};
    
    \node (model) [draw, rectangle, minimum width=2.5cm, minimum height=1cm, text centered, text width=2.5cm, right=of features, fill=white!10] {5. Feature Selection $\chi^2$};
    
    \node (train) [draw, rectangle, minimum width=2.5cm, minimum height=1cm, text centered, text width=2.5cm, right=of preprocess, fill=white!10] {6. Training XGBoost};
    
    \node (evaluate) [draw, rectangle, minimum width=2.5cm, minimum height=1cm, text centered, text width=2.5cm, right=of data, fill=white!10] {7. Evaluation};
    
    \node (results) [draw, rectangle, rounded corners, minimum width=2.5cm, minimum height=1cm, text centered, text width=2.5cm, right=of start, fill=white!10] {8. Results Analysis};

    \draw [-Stealth, thick, magenta] (start) -- (data);
    \draw [-Stealth, thick, magenta] (data) -- (preprocess);
    \draw [-Stealth, thick, magenta] (preprocess) -- (features);
    \draw [-Stealth, thick, magenta] (features) -- (model);
    \draw [-Stealth, thick, magenta] (model) -- (train);
    \draw [-Stealth, thick, magenta] (train) -- (model);
    \draw [-Stealth, thick, magenta] (train) -- (evaluate);
    \draw [-Stealth, thick, magenta] (evaluate) -- (results);
    \draw [-Stealth, thick, magenta] (evaluate) to[out=270,in=90] (train);
    \end{tikzpicture}
    \caption{Research process diagram.}
    \label{fig:research_process}
\end{figure}
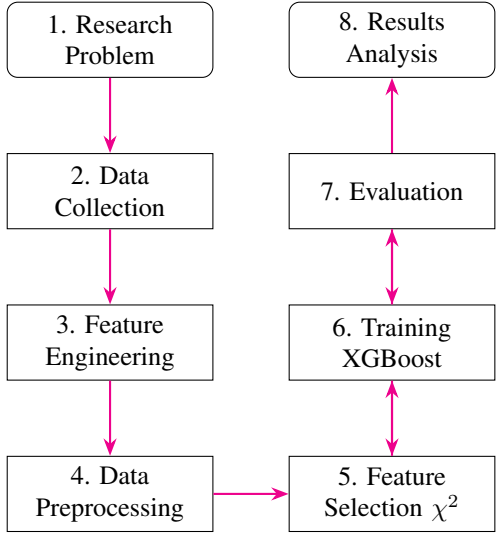

Figure \ref{fig:research_process} illustrates the research process, starting with the identification of the research problem, followed by data collection and preprocessing. After data preprocessing, feature engineering is performed, followed by feature selection using the chi-squared (\(\chi^2\)) test.
The selected features are used to train the model, which is then evaluated. The evaluation results may lead to retraining the model based on feedback. Finally, the results are analyzed and conclusions drawn.

Figure \ref{fig:research_process} illustrates the research process, starting with the identification of the research problem, followed by data collection and feature engineering. After feature engineering, data preprocessing is performed, followed by feature selection using the chi-squared (\(\chi^2\)) test.
The selected features are used to train the model, which is then evaluated. The evaluation results may lead to retraining the model based on feedback. Finally, the results are analyzed and conclusions drawn.

This paper makes several contributions to the field of cryptocurrency price prediction, outlined as follows:
\begin{itemize}
    \item We propose a robust machine learning strategy utilizing the XGBoost model, which generates accurate buy and sell signals. We fine-tune the XGBoost model across various parameters to select the most optimal accuracy while addressing potential overfitting through regularization techniques. Additionally, we compare the performance of this model with a fine-tuned Logistic Regression model, highlighting XGBoost's advantages in terms of classification accuracy and robustness. 
    \item We identify and analyze critical features encompassing various technical indicators and historical data, such as RSI, MACD, MOM, \%K, \%D, and CCI.
    \item We employ the chi-squared (\(\chi^2\)) statistical test to select the most important features that significantly contribute to the model's accuracy. Our analysis highlights the top 8 features: RSI30, MACD, MOM30, \%D30, \%D200, \%K200, \%K30, and RSI14.
    \item We demonstrate the balanced nature of our dataset with an approximately equal number of buy and sell signals, validating the use of accuracy as a reliable performance metric for our model.
    \item We provide a detailed comparison and discussion of the feature importance scores, emphasizing the critical role of selected features in predicting Bitcoin prices.
\end{itemize}

The rest of the paper is organized as follows: Section \ref{sec:related_work} presents related work. Section \ref{sec:data_collection_and_preprocessing_feature_engineering} explains how we collected and prepared the data and describes the feature engineering process. Section \ref{sec:Mathematical_Model} proposes the machine learning model and its mathematical formulation. In Section \ref{sec:results_and_evaluation}, we assess and compare our proposed model with existing ones. Section \ref{sec:discussion_and_comparison} compares this paper with existing works. Finally, Section \ref{sec:conclusion} concludes the paper.

\section{Related Work} \label{sec:related_work}

The rapid evolution and growth of the cryptocurrency market, characterized by high volatility and lack of regulation, poses significant challenges and opportunities for developing effective trading strategies. Despite its growing importance, a limited number of studies have focused on building reliable/profitable trading strategies within this market. This section reviews and summarizes the current literature, emphasizing key advancements and methodologies that utilize machine learning models for forecasting and trading, both in traditional financial assets (e.g., stocks \cite{zhao2023stock}) and cryptocurrencies (e.g., Bitcoin \cite{nakamoto2008bitcoin} and Ethereum \cite{buterin2014ethereum}).

In 2017, Shynkevich et al. \cite{shynkevich2017forecasting} asserted that developing an accurate predictive system to forecast future changes in stock prices is vital for investment management and algorithmic trading. Their study investigates how the combination of forecast horizon and input window length for variable horizon forecasting affects the predictive system's performance. Machine learning algorithms utilize technical indicators as input features to forecast future movements in stock prices. The dataset comprises ten years of daily price time series for fifty stocks. The optimal prediction performance is observed when the input window length is approximately equal to the forecast horizon. This distinct pattern is analyzed using multiple performance metrics, including prediction accuracy, winning rate, return per trade, and Sharpe ratio.

Lin et al. \cite{lin2020forecasting} presented an advanced trading strategy using a Recurrent Neural Network (RNN) to forecast stock prices, specifically focusing on the opening and closing prices and their differences for the S\&P 500 and Dow Jones indices. By incorporating data pre-processing techniques such as the normalized first-order difference method and zero-crossing rate (ZCR), the model enhances the accuracy of its predictions. Results show that this approach performs better than previous methods, with metrics including a Root Mean Squared Error (RMSE) of 12.933, a Mean Absolute Error (MAE) of 10.44, and an accuracy of 59.4\%. The study concludes that the RNN-based method provides crucial/valuable insights for market strategies, with plans to integrate Natural Language Processing (NLP) and statistical methods to improve the model's performance further \cite{lin2020forecasting}.

Liu et al. \cite{liu2021forecasting} constructed a feature system with 40 determinants affecting Bitcoin's price, considering factors such as the cryptocurrency market (e.g., trading volume), public attention, and the macroeconomic environment (e.g., the exchange rate of the US dollar). The authors applied the stacked denoising autoencoders (SDAE) deep learning method for price prediction. Comparative analysis with established methods like backpropagation neural network (BPNN) and support vector regression (SVR) revealed superior performance by the SDAE model in directional and level predictions. Evaluation metrics, including mean absolute percentage error (MAPE), RMSE, and directional accuracy (DA), consistently favored the SDAE model. This underscores the efficacy of the SDAE approach in forecasting Bitcoin prices, surpassing widely recognized benchmark methods \cite{liu2021forecasting}.

Passalis et al. \cite{passalis2018temporal} introduced a novel temporal-aware neural bag-of-features (BoF) model adapted for time-series forecasting using high-frequency limit order book data. Limit order book data, which provides comprehensive insights into stock behavior, presents challenges such as high dimensionality and large volume. The proposed temporal BoF model addresses these challenges by utilizing radial basis function (RBF) and accumulation layers to capture both short-term and long-term dynamics in the data. This enhances the model's ability to forecast complex temporal patterns. The model's effectiveness is validated with a large-scale dataset containing over 4.5 million limit orders, demonstrating superior performance compared to other methods. Key metrics achieved include a precision of 46.01, recall of 56.21, F1 score of 45.46, and Cohen's $\kappa$ of 0.2222 for a short-term horizon of 10 (as shown in Table 3 of \cite{passalis2018temporal}). Future research directions include extending the method for interactive exploratory analysis and improving its robustness to distribution shifts through multi-scale training schemes and integration with recurrent models to enhance its dynamic modeling capacity \cite{passalis2018temporal}.

Jaquart et al. \cite{jaquart2022machine} utilized a variety of machine learning models to forecast and trade in the daily cryptocurrency market. These models are trained to predict whether the market movements of the top 100 cryptocurrencies will rise or fall daily. Data streaming is facilitated from CoinGecko \cite{coingecko}. The findings indicate that all models yield statistically significant predictions, with average accuracy rates ranging from 52.9\% to 54.1\% across all cryptocurrencies. Notably, when focusing on predictions with the highest 10\% model confidences per class and day, accuracy rates increase from 57.5\% to 59.5\%.

Furthermore, the authors observe that employing a long-short portfolio strategy based on predictions from ensemble models utilizing LSTM and Gated Recurrent Unit (GRU) architectures results in annualized out-of-sample Sharpe ratios of 3.23 and 3.12, respectively, after accounting for transaction costs. In contrast, a buy-and-hold benchmark market portfolio strategy only yields a Sharpe ratio (SR) of 1.33. These findings suggest a departure from weak-form efficiency in the cryptocurrency market, although the impact of certain arbitrage constraints remains a consideration \cite{jaquart2022machine}.

Saad et al. \cite{saad2019toward} delved into the network dynamics of cryptocurrencies to understand the factors driving their price hikes. Analyzing user and network activity, they identified key indicators influencing price fluctuations. Their approach integrates economic theories with machine learning methods to construct accurate price prediction models for Bitcoin \cite{nakamoto2008bitcoin} and Ethereum \cite{wood2014ethereum}. Experimental results using large datasets validated their models, achieving up to 99\% accuracy in price prediction. Their research contributes insights into the network features shaping cryptocurrency prices and advances predictive modeling in this domain.

Akyildirim et al. \cite{akyildirim2021prediction} analyzed the ability to predict price movements of the twelve most traded cryptocurrencies. The authors utilized machine learning algorithms such as SVMs (Support Vector Machines), logistic regression, ANN (Artificial Neural Networks), and random forests. These algorithms leverage historical price data and technical indicators as features to forecast/predict future prices. On average, all four algorithms achieve classification accuracy of over 50\% for all cryptocurrencies and periods. This suggests that there is some predictability in cryptocurrency price trends. At both daily and minute intervals, machine learning algorithms achieve predictive accuracies ranging from 55\% to 65\%. Among these, SVMs consistently demonstrate superior predictive accuracy compared to other algorithms, including logistic regression, ANN, and random forests \cite{akyildirim2021prediction}.

Recently, Hafid et al. \cite{hafid2023bitcoin} introduced a machine learning classification approach to predict market trends, discerning their upward or downward movement. The authors identify key features, incorporating historical data such as volume, and utilize technical indicators such as the Relative Strength Index (RSI), Moving Average Convergence Divergence (MACD), and Exponential Moving Average (EMA) as inputs for the Random Forest classifier. The proposed approach is validated by comprehensively analyzing Bitcoin's closing prices. Data streaming is facilitated from Binance via the Binance API \cite{binance_data}. The authors employ various simulations during the evaluation, including a backtesting strategy. The results demonstrate that their machine learning method effectively signals optimal moments for buying or selling, achieving an accuracy of 86\%.

In a more recent study, Roy et al. \cite{roy2023forecasting} utilized a Yahoo finance dataset from 2016 to 2021 to train an LSTM model, achieving impressive metrics: MAE of 253.30, RMSE of 409.41, and $R^{2}$ of 0.9987 (as shown in Table 2 of \cite{roy2023forecasting}). This model outperformed Bi-LSTM and GRU variants, demonstrating its robustness in minimizing prediction errors. The findings highlight the potential of LSTM models in providing reliable buy and sell signals, thereby assisting investors in making informed decisions in the volatile cryptocurrency market. The study also suggests future enhancements by incorporating advanced machine learning techniques and diverse data sources such as sentiment analysis, which could further improve prediction accuracy and model interpretability \cite{roy2023forecasting}.

\section{Data Collection, Preprocessing and Feature Engineering}
\label{sec:data_collection_and_preprocessing_feature_engineering}
In this section, we describe the data collection, preprocessing, as well as feature engineering process. The list of our abbreviations and notations is provided in Table \ref{tab:notation}.


\begin{table*}[ht!]
\renewcommand{\arraystretch}{1.2}
\caption{Nomenclature and Definitions.}
\label{tab:notation}
\centering
\footnotesize
\begin{tabular}{lp{0.75\linewidth}}
\hline
\textbf{Notation} & \textbf{Description} \\
\hline
\(m\) & Total number of samples or observations\\
\(m_{\text{train}}\) & Number of training samples \\
\(m_{\text{test}}\) & Number of testing samples \\
$n$ & Number of features \\
$\mathcal{C}_p^{(i)}$ & Closing price of the current period \(t_i\) \\
$\mathcal{O}_p^{(i)}$ & Opening price of the current period \(t_i\) \\
$\mathcal{H}_p^{(i)}$ & Highest price of the current period \(t_i\) \\
$\mathcal{L}_p^{(i)}$ & Lowest price of the current period \(t_i\) \\
$\mathcal{V}^{(i)}$ & Volume of the cryptocurrency being traded at time $t_i$ \\
$\text{QAV}^{(i)}$ & Total trading value at time $t_i$ \\
$\text{NOT}^{(i)}$ & Number of trades at time $t_i$ \\
$\text{TBBV}^{(i)}$ & Total volume of Bitcoin bought at time $t_i$ \\
$\text{RSI}_{\tau}^{(i)}$ & Relative strength index at time $t_i$ within a time period $\tau$ \\
$\text{MACD}^{(i)}$ & Moving average convergence divergence at time $t_i$ \\
$\text{EMA}_{\tau}^{(i)}$ & Exponential moving average at time $t_i$ within a period of time $\tau$ \\
$\text{PROC}_{\tau}^{(i)}$ & Price rate of change at time $t_i$ within a period of time $\tau$ \\
$ \%K_{\tau}^{(i)}$ & Stochastic oscillator at time $t_i$ within a period of time $\tau$ \\
$ \%D_{\tau}^{(i)}$ & Smoothed stochastic oscillator at time $t_i$ within a period of time $\tau$ \\
$\text{MOM}_{\tau}^{(i)}$ & Momentum at time $t_i$ within a period of time $\tau$ \\
$\text{BB}_{\tau}^{(i)}$ & Bollinger Bands at time $t_i$ within a period of time $\tau$ \\
$\text{ATR}_{\tau}^{(i)}$ & Average True Range at time $t_i$ within a period of time $\tau$ \\
$\text{CCI}_{\tau}^{(i)}$ & Commodity Channel Index at time $t_i$ within a period of time $\tau$ \\
$\text{\%R}_{\tau}^{(i)}$ & Williams \%R indicator value at time $t_i$ within a period of time $\tau$ \\
$\text{CMF}_{\tau}^{(i)}$ & Chaikin Money Flow at time $t_i$ within a period of time $\tau$ \\
$\text{OBV}^{(i)}$ & On-balance Volume at time $t_i$ \\
$\text{ADL}^{(i)}$ & Accumulation/Distribution line at time $t_i$ \\
$K$           & Total number of trees in the ensemble \\
$\eta$ & Learning rate \\
$\lambda$, $\alpha$ & Regularization parameters \\
$N$ & Number of trees \\
$\Delta t$ & Time interval \\
$s$ & Span ($s \geq 1 $)\\
\hline
\end{tabular}
\normalsize
\end{table*}


\subsection{Data Collection} \label{subsec: data collection and preprocessing}

We obtain Bitcoin historical market data from Binance via Binance API \cite{binance_data}. The dataset covers the period from February 1, 2021, to February 1, 2022, with a time interval of 15 minutes (i.e., $\Delta t$ = $15$ minutes). We chose $15$ minutes because it gives us good accuracy. We split the data into $80\%$ for the training set and $20\%$ for the testing set. 

We choose this shorter time interval for several reasons, including the high volatility of the Bitcoin market. In highly volatile markets, shorter intervals are often preferred to capture rapid changes, whereas less volatile markets may be suitable for longer intervals. Additionally, our choice aligns with the time intervals used in existing works for meaningful comparisons.

\begin{figure}[!t]
   \centering
   \includegraphics[width=3.5in]{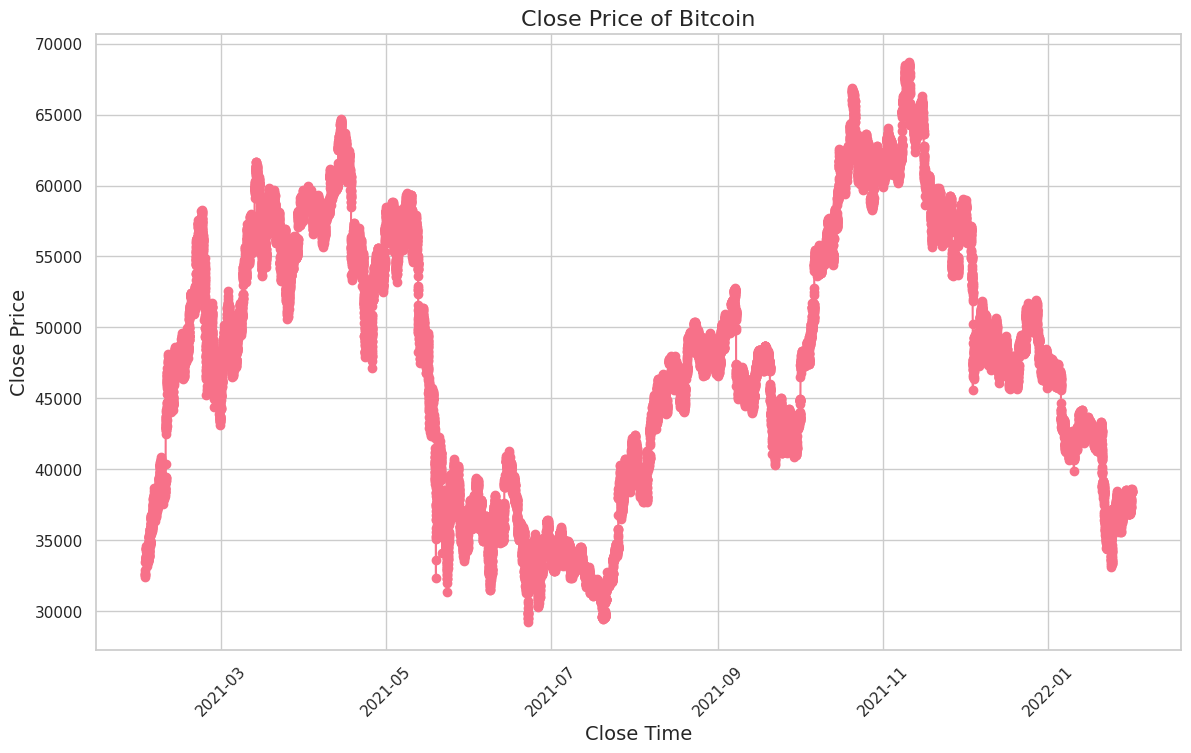}
   \caption{Close price of Bitcoin from February 1, 2021, to February 1, 2022, showing trends and fluctuations over the selected timeframe.}
   \label{fig:close_price_Bitcoin}
\end{figure}

Figure \ref{fig:close_price_Bitcoin} shows the trend of Bitcoin's closing prices from February 1, 2021, to February 1, 2022, with a time interval of 15 minutes. The x-axis represents the ''Close Time," indicating the specific times at which the Bitcoin prices were recorded. In contrast, the y-axis represents the ''Close Price" regarding Bitcoin's value at those respective times. Each data point on the plot is marked with a circular marker, helping to visualize the fluctuations and trends in Bitcoin prices. This figure clearly represents the volatility and trends in Bitcoin's market value over the selected timeframe.

\subsection{Feature Engineering and Preprocessing}
\label{sec:Feature_Engineering_preprocessing}

This section illustrates the various features incorporated in this case study. Specifically, we employ historical market data alongside technical indicators. Additionally, we preprocess our dataset.

\subsubsection{Historical Data}
In the context of historical data analysis, we focus on the closing price and volume.  
\subsubsection{Technical Indicators}
Technical analysis indicators represent a trading discipline utilized to assess investments and pinpoint trading opportunities by analyzing statistical trends derived from trading activities, including price movements and volume \cite{murphy1999technical}.
In this study, we incorporate various technical indicators such as the exponential moving average, moving average convergence divergence, relative strength index, momentum, price rate of change, and stochastic oscillator (all these technical indicators are well described and mathematically formalized in \cite{hafid2023bitcoin}). 

Additionally, we implement further technical indicators, including Bollinger Bands, Average True Range, Commodity Channel Index, Williams \%R, and Chaikin Money Flow.

\subsection*{Bollinger Bands}
Bollinger Bands (BB) are a volatility indicator that consists of a moving average (MA) and two standard deviation lines (BB\_up and BB\_dn) above and below the moving average. They help identify overbought and oversold conditions \cite{bollinger2002bollinger}. This indicator can be formulated as follows:

\begin{align*}
\text{MA}_{\tau}^{(i)} &= \frac{1}{\tau} \sum_{j=0}^{\tau-1} \mathcal{C}_p^{(j)} \\
\text{BB}_{\text{up}, \tau}^{(i)} &= \text{MA}_{\tau}^{(i)} + 2 \times \sigma(\mathcal{C}_p^{(i)}) \\
\text{BB}_{\text{dn}, \tau}^{(i)} &= \text{MA}_{\tau}^{(i)} - 2 \times \sigma(\mathcal{C}_p^{(i)})
\end{align*}

Where $\sigma(\mathcal{C}_p^{(i)})$ is the standard deviation of the closing prices over the period \(\tau\) (\(\tau = 20\) in this case study).

\subsection*{Average True Range}
The Average True Range  (ATR) is a measure of market volatility. It calculates the average of the true range over a specified period (e.g., 14 days) \cite{wilder1978new}. The mathematical formula for the \(\text{ATR}\) indicator is given by:

First, calculate the True Range (\(\text{TR}\)) for each period \(t\):

\[
\text{TR}^{(i)} = \max(\mathcal{H}_p^{(i)} - \mathcal{L}_p^{(i)}, |\mathcal{H}_p^{(i)} - \mathcal{C}_p^{(i-1)}|, |\mathcal{L}_p^{(i)} - \mathcal{C}_p^{(i-1)}|)
\]

Then, calculate the (\(\text{ATR}\)) over a specified number of periods \(\tau\):

\[
\text{ATR}_{\tau}^{(i)} = \frac{1}{\tau} \sum_{i=1}^{\tau} \text{TR}^{(i)}
\]

\subsection*{Commodity Channel Index}
The Commodity Channel Index (CCI) measures a security's price variation from its average price. High positive values indicate overbought conditions, while low negative values indicate oversold conditions \cite{lambert1983commodity}. This indicator is defined as follows:
\begin{align*}
    \text{TP} &= \frac{\mathcal{H}_p^{(i)} + \mathcal{L}_p^{(i)} + \mathcal{C}_p^{(i)}}{3} \\
    \text{SMA}_{\tau}(\text{TP}) &= \frac{1}{\tau} \sum_{i=0}^{\tau-1} \text{TP}_i \\
    \text{MAD}_{\tau} &= \frac{1}{\tau} \sum_{i=0}^{\tau-1} |\text{TP}_i - \text{SMA}{\tau}(\text{TP})| \\
    \text{CCI}_{\tau}^{(i)} &= \frac{\text{TP} - \text{SMA}_{\tau}(\text{TP})}{0.015 \times \text{MAD}_{\tau}}
\end{align*}

Where \(\text{TP}\) denotes the Typical Price, computed as the arithmetic mean of the High, Low, and Close prices; \(\text{SMA}_{\tau}(\text{TP})\) represents the Simple Moving Average of the Typical Price over a specified period \(\tau\), and \(\text{MAD}_{\tau}\) signifies the Mean Absolute Deviation of the Typical Price over the same period \(\tau\).

\subsection*{Williams \%R}
The Williams \%R is a momentum indicator that measures the level of the close relative to the highest high for the look-back period. It helps to identify overbought and oversold conditions \cite{williams1979made}. This indicator can be formulated as follows:

\begin{align*}
    \%R &= -100 \times \frac{\text{Highest High}_\tau - \text{Close}}{\text{Highest High}_\tau - \text{Lowest Low}_\tau}
\end{align*}

\begin{align*}
    \text{\%R}_{\tau}^{(i)} &= -100 \times \frac{\text{Highest High}_{\tau} - \mathcal{C}_p^{(i)}}{\text{Highest High}_{\tau} - \text{Lowest Low}_{\tau}}
\end{align*}
where:
\begin{itemize}
    \item \(\text{Highest High}_n\) is the highest high over the past \(n\) periods.
    \item \(\text{Lowest Low}_n\) is the lowest low over the past \(n\) periods.
    \item \(\text{Close}\) is the closing price of the current period.
\end{itemize}
\subsection*{Chaikin Money Flow}
The Chaikin Money Flow (CMF) is an indicator that measures the accumulation and distribution of a security over a specified period. It is calculated using the Money Flow Multiplier and the Money Flow Volume over the period. This indicator is defined as follows:

\begin{align*}
    \text{MFV}^{(i)} &= \frac{(\mathcal{C}_p^{(i)} - \mathcal{L}_p^{(i)}) - (\mathcal{H}_p^{(i)} - \mathcal{C}_p^{(i)})}{\mathcal{H}_p^{(i)} - \mathcal{L}_p^{(i)}} \times \mathcal{V}^{(i)} \\
    \text{CMF}_\tau &= \frac{\sum_{t=1}^{n} \text{MFV}^{(i)}}{\sum_{t=1}^{n} \mathcal{V}^{(i)}}
\end{align*}

\subsection*{On-balance Volume}
The on-balance volume (OBV) is a technical trading momentum indicator that uses volume flow to predict changes in stock price. The OBV is calculated by adding volume on up days and subtracting it on down days. The cumulative OBV line then provides insight into the strength of a trend \cite{achelis2001technical}. This indicator can be defined as follows:

\[
\text{OBV}^{(i)} = \text{OBV}^{(i-1)} + 
\begin{cases} 
\mathcal{V}^{(i)} & \text{if } \mathcal{C}_p^{(i)} > \mathcal{C}_p^{(i-1)} \\
-\mathcal{V}^{(i)} & \text{if } \mathcal{C}_p^{(i)} < \mathcal{C}_p^{(i-1)} \\
0 & \text{if } \mathcal{C}_p^{(i)} = \mathcal{C}_p^{(i-1)}
\end{cases}
\]

\subsection*{Accumulation/Distribution Line}
The Accumulation/Distribution (A/D) line is a technical indicator that uses volume and price to assess the cumulative flow of money into and out of a security. It measures the supply and demand by evaluating whether investors are accumulating (buying) or distributing (selling) a particular stock \cite{murphy1999technical}. This indicator can be expressed as follows:
\[
\text{ADL}^{(i)} = \text{ADL}^{(i-1)} + \left( \frac{(\mathcal{C}_p^{(i)} - \mathcal{L}_p^{(i)}) - (\mathcal{H}_p^{(i)} - \mathcal{C}_p^{(i)})}{\mathcal{H}_p^{(i)} - \mathcal{L}_p^{(i)}} \right) \times \mathcal{V}^{(i)}
\]

\subsection*{Signal}
Let $\mathcal{Y}$ be a random variable that takes the values of 1 or -1 (\textit{Buy} and \textit{Sell}, respectively).

To generate \textit{Buy} and \textit{Sell} signals, we employ a technical indicator called Moving Average (MA). MA identifies the trend of the market. The MA rule that generates \textit{Buy} and \textit{Sell} signals at time $t$ consists of comparing two moving averages. Formally, the rule is expressed as follows:

\begin{equation}
\mathcal{Y}(t_i) = \begin{cases} 1 & \mbox{if } \text{MA}_{s,t_i} \geq \text{MA}_{l,t_i} , \\ 
                              -1 & \mbox{if } \text{MA}_{s,t_i} < \text{MA}_{l,t_i}\end{cases}
\end{equation}
where 
\begin{equation}
\text{MA}_{j,t_{i}} = (1/j) \sum_{k=0}^{j-1} \mathcal{C}_p^{(i-k)}, \hspace{0.25 cm} \text{for} \hspace{0.15 cm} j= s \hspace{0.15 cm} \text{or} \hspace{0.15 cm} l;
\end{equation}

\textit{s} (\textit{l}) is the length of the short (long) MA ( $\textit{s} < \textit{l}$). We denote the
MA indicator with MA lengths \textit{s} and \textit{l} by MA(\textit{s}, \textit{l}). In this paper, we consider the MA(10, 60) because of it high accuracy.

\begin{figure}[!t]
   \centering
   \includegraphics[width=3.5in]{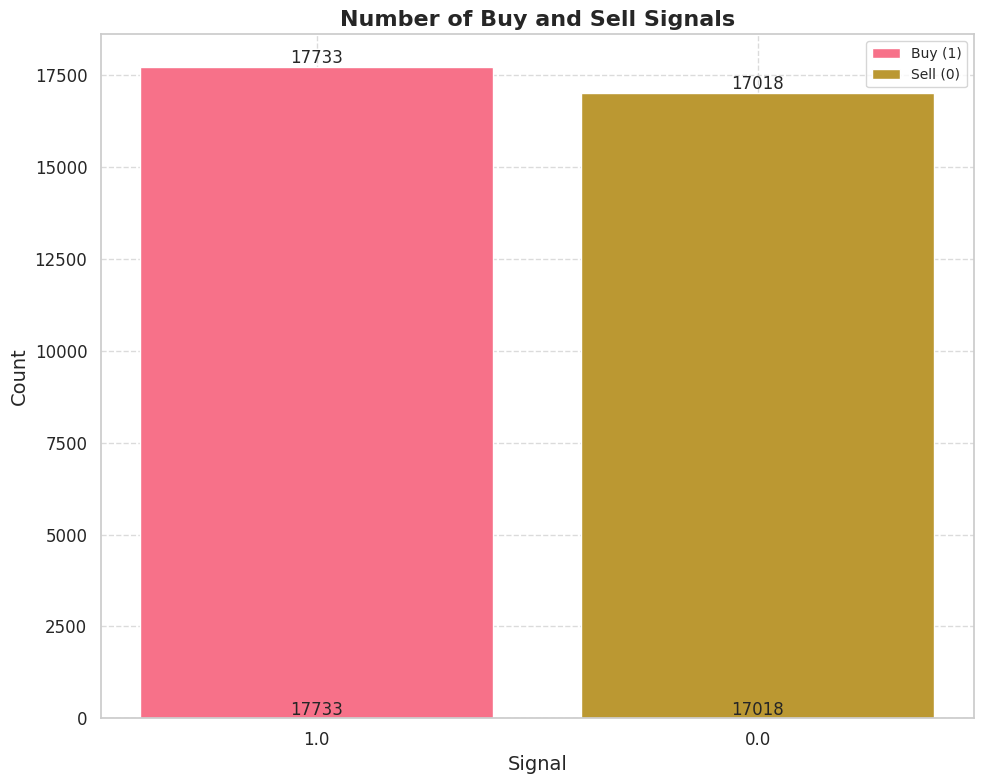}
   \caption{Number of Buy and Sell signals.}
   \label{fig_count}
\end{figure}

Figure \ref{fig_count} shows the close price chart indicating the number of Buy and Sell signals. The approximately equal distribution of Buy and Sell signals benefits classification machine learning problems, implying a balanced dataset. Consequently, accuracy can be considered a reliable measure for evaluating the model's performance.
\subsubsection{Data Preprocessing}
We scale the data by using the \texttt{StandardScaler} from \texttt{sklearn.preprocessing} module. Let's denote the elements of the matrix \(X\) as \(x_{ij}\), where \(i\) represents the row index (sample) and \(j\) represents the column index (feature). The transformation applied by the \texttt{StandardScaler} to each feature \(j\) is as follows: 1) Compute the mean (\(\mu_j =\frac{1}{m} \sum_{i=1}^{m} x_{ij}\)) and standard deviation (\(\sigma_j =\sqrt{\frac{1}{m} \sum_{i=1}^{m} (x_{ij} - \mu_j)^2}\)) of feature \(j\), where \(m\) is the number of samples (rows), \(x_{ij}\) is the element at the \(i\)-th row and \(j\)-th column of \(X\). 2) Apply the transformation to each element of feature \(j\): \[x'_{ij} = \frac{x_{ij} - \mu_j}{\sigma_j}\] where \(x'_{ij}\) is the scaled value of \(x_{ij}\).

\subsubsection{Data Preprocessing}
We scale the data using the \texttt{StandardScaler} from the \texttt{sklearn.preprocessing} module. Let's denote the elements of the matrix \(X\) as \(x_{ij}\), where \(i\) represents the row index (sample) and \(j\) represents the column index (feature). The transformation applied by the \texttt{StandardScaler} to each feature \(j\) is as follows: 
1) Compute the mean (\(\mu_j =\frac{1}{m} \sum_{i=1}^{m} x_{ij}\)) and standard deviation (\(\sigma_j =\sqrt{\frac{1}{m} \sum_{i=1}^{m} (x_{ij} - \mu_j)^2}\)) of feature \(j\), where \(m\) is the number of samples (rows). 
2) Apply the transformation to each element of feature \(j\):
\[
x'_{ij} = \frac{x_{ij} - \mu_j}{\sigma_j}
\]
where \(x'_{ij}\) is the scaled value of \(x_{ij}\).

In addition to scaling, the data is split into training and testing sets using a time-series approach. Since the dataset is sequential, a percentage-based split is applied to maintain the temporal order. This ensures that the model is tested on future data, preventing data leakage from the training set. The mathematical formulation for the splitting process is provided in Appendix \ref{appendix_A}.

\section{Mathematical Model}
\label{sec:Mathematical_Model}

Let $(x^{(i)}, y^{(i)})$ denotes a single sample/observation, and the set of samples is represented by:  

\begin{equation}
\mathcal{S} = \left\{(x^{(1)}, y^{(1)}), (x^{(2)}, y^{(2)}), \dots, (x^{(m)}, y^{(m)}) \right\}
\end{equation}
where \(x^{(i)} \in \mathbb{R}^{n}\) and \(y^{(i)} \in \mathcal{T}\).

Considering both technical indicators and historical data for price prediction necessitates the integration of diverse datasets. To achieve this, we combine technical indicators and historical data as inputs to our model. The feature vector at a given time $t$ can be expressed as follows:

\begin{equation} 
\label{equation:feature}
\mathbf{x_{\text{org}}}^{(i)} =
\begin{bmatrix}
\mathcal{C}_p^{(i)} \\
\mathcal{V}^{(i)} \\
\text{RSI}_{14}^{(i)} \\
\text{RSI}_{30}^{(i)} \\
\text{RSI}_{200}^{(i)} \\
\text{MOM}_{10}^{(i)} \\
\text{MOM}_{30}^{(i)} \\
\text{MACD}^{(i)} \\
\text{PROC}_{9}^{(i)} \\
\text{EMA}_{10}^{(i)} \\
\text{EMA}_{30}^{(i)} \\
\text{EMA}_{200}^{(i)} \\
\%K_{10}^{(i)} \\
\%K_{30}^{(i)} \\
\%K_{200}^{(i)} \\
\end{bmatrix}
, \quad \mathbf{x}^{(i)} \in \mathbb{R}^{n}
\end{equation}

Let's extend the generality of our model by stacking all feature vectors into a matrix $\mathbf{X_{\text{org}}}$. This matrix can now be expressed as follows:

\begin{equation}
\mathbf{X_{\text{org}}} =
\begin{bNiceMatrix}[first-row,
                    first-col,
                    last-col,
                    nullify-dots,
                    code-for-first-row = \color{magenta},
                    code-for-first-col = \color{blue}
                    ]
& \mathcal{C}_p & \mathcal{V} &  \cdots & \%K_{200} & \\
  & x_{11} & x_{12} & \cdots &  x_{1n} &  \\
 & x_{21} & x_{22}  & \cdots & x_{2n} & \\
 & x_{31} & x_{32} & \cdots & x_{3n} &  \\
 & \vdots  & \vdots & \cdots  & \vdots & \\
 & x_{m-11} & x_{m-12} & \cdots & x_{m-1n} & \\
 & x_{m1} & x_{m2} & \cdots & x_{mn} & \\
\end{bNiceMatrix}
\end{equation}

Where:
\begin{align*} 
\mathcal{C}_{p} &= \begin{bmatrix} \mathcal{C}_{p}^{(1)} \\ \mathcal{C}_{p}^{(2)} \\ \vdots \\ \mathcal{C}_{p}^{(m)} \end{bmatrix},
\mathcal{V} &= \begin{bmatrix} \mathcal{V}^{(1)} \\ \mathcal{V}^{(2)} \\ \vdots \\ \mathcal{V}^{(m)} \end{bmatrix},
\hdots,
\%K_{200}  &= \begin{bmatrix} \%K_{200}^{(1)} \\ \%K_{200}^{(2)} \\ \vdots \\ \%K_{200}^{(m)} \end{bmatrix}
\end{align*}

The output matrix can be expressed as follows:
\begin{equation*}
    Y = \begin{bmatrix} y^{(1)} \\ y^{(2)} \\ \vdots \\ y^{(m)} \end{bmatrix}
\end{equation*}

where $y^{(i)}$ belongs to the set $\mathcal{T}$, and $i$ ranges from 1 to $m$.

For feature selection, We import the \texttt{SelectKBest} class from the \texttt{feature\_selection} module in scikit-learn. \texttt{SelectKBest} is a feature selection method that selects the top \(k\) features based on a scoring function, which in this case is \texttt{chi2}. The \texttt{chi2} function is a scoring function used to evaluate the importance of features based on the chi-squared (\(\chi^2\)) statistical test. This test is typically used for categorical data to measure the independence between each feature and the target variable.

The \(\chi^2\) test statistic is calculated as follows:

\[
\chi^2 = \sum \frac{(O_i - E_i)^2}{E_i}
\]
where:
\begin{itemize}
    \item \(O_i\) is the observed frequency (actual count) of class \(i\).
    \item \(E_i\) is the expected frequency (theoretical count) of class \(i\) if there were no association between the feature and the target variable.
\end{itemize}

We select the $k=8$ top features with high scores based on this feature selection method. The feature vector $\mathbf{x'}^{(i)}$ will be replaced by $\mathbf{x}^{(i)}$ which can be expressed as follows:

\begin{equation} 
\label{equation:feature_new}
\mathbf{x}^{(i)} =
\begin{bmatrix}
\text{RSI}_{30}^{(i)} \\
\text{MACD}^{(i)} \\
\text{MOM}_{30}^{(i)} \\
\%D_{30}^{(i)} \\
\%D_{2000}^{(i)} \\
\%K_{200}^{(i)} \\
\%K_{30}^{(i)} \\
\text{RSI}_{14}^{(i)} \\
\end{bmatrix}
, \quad \mathbf{x}^{(i)} \in \mathbb{R}^{n}
\end{equation}

The matrix $\mathbf{X}_{\text{org}}$ will be replaced by $\mathbf{X}$ as depicted in Figure \ref{fig:feature-matrix}.

\begin{figure*}[!ht]
    \begin{equation}
    \mathbf{X} =
    \begin{bNiceMatrix}[first-row,
                        first-col,
                        nullify-dots,
                        code-for-first-row = \color{magenta},
                        code-for-first-col = \color{blue}
                        ]
    & \text{RSI}_{30} & \text{MACD} & \text{MOM}_{30} & \%D_{30} & \%K_{200} & \%K_{200} & \%K_{30} & \text{RSI}_{14} \\
    & \text{RSI}_{30}^{(1)} & \text{MACD}^{(1)} & \text{MOM}_{30}^{(1)} & \%D_{30}^{(1)} & \%K_{200}^{(1)} & \%K_{200}^{(1)} & \%K_{30}^{(1)} & \text{RSI}_{14}^{(1)} \\
    & \text{RSI}_{30}^{(2)} & \text{MACD}^{(2)} & \text{MOM}_{30}^{(2)} & \%D_{30}^{(2)} & \%K_{200}^{(2)} & \%K_{200}^{(2)} & \%K_{30}^{(2)} & \text{RSI}_{14}^{(2)} \\
    & \text{RSI}_{30}^{(3)} & \text{MACD}^{(3)} & \text{MOM}_{30}^{(3)} & \%D_{30}^{(3)} & \%K_{200}^{(3)} & \%K_{200}^{(3)} & \%K_{30}^{(3)} & \text{RSI}_{14}^{(3)} \\
    & \vdots & \vdots & \vdots & \vdots & \vdots & \vdots & \vdots & \vdots \\
    & \text{RSI}_{30}^{(m)} & \text{MACD}^{(m)} & \text{MOM}_{30}^{(m)} & \%D_{30}^{(m)} & \%K_{200}^{(m)} & \%K_{200}^{(m)} & \%K_{30}^{(m)} & \text{RSI}_{14}^{(m)} \\
    \end{bNiceMatrix}
    \end{equation}
    \caption{Matrix $\mathbf{X}$ showing various technical indicators (features) selected by the \(\chi^2\) statistical test. Each row represents different instances or observations, while each column corresponds to a specific technical indicator or feature. The notation $\text{RSI}_{30}^{(i)}$, $\text{MACD}^{(i)}$, $\text{MOM}_{30}^{(i)}$, etc., denote different instances/observations of the respective technical indicators.}
    \label{fig:feature-matrix}
\end{figure*}

In this case study, the problem is to minimize the cost function for XGBoost, which is a regularized finite-sum minimization problem defined as:
\begin{equation}
\min_{\Theta} J(\Theta) := \sum_{i=1}^{m_{\text{train}}} L(y_i, \hat{y}_i) + \sum_{k=1}^{K} \mathcal{R}(f_k)
\end{equation}

Where:
\begin{itemize}
   \item \( \Theta \) denotes the set of parameters to be optimized during the training process. The parameter set \( \Theta \) is defined as follows:
   \[
    \Theta = \begin{pmatrix}
    C & \gamma & \eta & D_{\text{max}} & W_{\text{min}} & N & \alpha & \lambda & S
    \end{pmatrix}
  \]

   \item \( L(y_i, \hat{y}_i) \) is the loss function that measures the difference between the true label \( y_i \) and the predicted label \( \hat{y}_i \) for the \( i \)-th instance. In the context of this case study, we employ the logistic loss function, which is expressed as follows:
  \begin{equation}
  L(y_i, \hat{y}_i) = -\left[y_i \log(\hat{y}_i) + (1 - y_i) \log(1 - \hat{y}_i)\right]
  \end{equation}
  Here, \( y_i \) is the true label (1 or 0), and \( \hat{y}_i \) is the predicted probability of class 1.

  \item \( \mathcal{R}(f_k) \) represents the regularization term for each tree to control its complexity. It typically includes both \( L_1 \) and \( L_2 \) regularization. Assuming \(T\) is the number of leaves in tree \(f_k\) and \(w_{j, k}\) is the weight for leaf \(j\) in tree \(f_k\), the regularization term for tree \(f_k\) is:
  \begin{equation}
      \mathcal{R}(f_k) = \gamma T + \frac{1}{2} \lambda \sum_{j=1}^T w_{j, k}^2 + \alpha \sum_{j=1}^T |w_{j, k}|
  \end{equation}
  The regularization terms (\( \mathcal{R}(f_k) \)) help control the complexity of individual trees in the ensemble, preventing overfitting.
\end{itemize}

During training, XGBoost aims to find the set of parameters (\( \Theta \)) that minimizes the overall cost function. The optimization is  performed using gradient boosting (Algorithm \ref{alg:gradient_boosting}), which involves iteratively adding weak learners to the ensemble to reduce the residual errors.

\begin{algorithm}[ht]
 \caption{Algorithm for XGBoost (Gradient Boosting for classification).} 
\begin{algorithmic}[1]
\STATE // Initialize model with constant value
\STATE Initialize model with constant value \(\hat{y}_i^{(0)}\).
\FOR{$k = 1$ to $K$}
    \STATE // Compute the gradient for the logistic loss function
    \STATE Compute the gradient \(g_i^{(k)} = \frac{\partial L(y_i, \hat{y}_i^{(k-1)})}{\partial \hat{y}_i^{(k-1)}}\).
    \STATE // Compute the Hessian for the logistic loss function
    \STATE Compute the Hessian \(h_i^{(k)} = \frac{\partial^2 L(y_i, \hat{y}_i^{(k-1)})}{\partial \hat{y}_i^{(k-1) 2}}\).
    \STATE // Fit a regression tree to the gradients and Hessians
    \STATE Fit a regression tree \(f_k\) to the gradients \(g_i^{(k)}\) and Hessians \(h_i^{(k)}\).
    \STATE // Compute the weight for each leaf in the tree
    \STATE Compute the weight for each leaf \(w_{j, k}\) using:
    \[
    w_{j, k} = -\frac{\sum_{i \in I_j} g_i^{(k)}}{\sum_{i \in I_j} h_i^{(k)} + \lambda}
    \]
    \STATE // Add the regularization term to control tree complexity
    \STATE Add the regularization term \(\mathcal{R}(f_k)\) to control tree complexity.
    \STATE // Update the prediction with the new tree
    \STATE Update the prediction:
    \[
    \hat{y}_i^{(k)} = \hat{y}_i^{(k-1)} + \eta f_k(x_i)
    \]
\ENDFOR
\STATE // Output the final model
\STATE Output the final model: \(\hat{y}_i = \sum_{k=1}^{K} \eta f_k(x_i)\).
\end{algorithmic}
\label{alg:gradient_boosting}
\end{algorithm}

\begin{table}[!t]
\centering
\caption{\textit{Parameter grid for GridSearchCV.}}
\label{tab:param_grid}
\begin{tabular}{ll}
\toprule
\textbf{Parameter} & \textbf{Values} \\
\midrule
\( N \) & 300, 400 \\
\( \eta \) & 0.01, 0.1, 0.2 \\
\( D_{\max} \) & 3, 4 \\
\( W_{\text{min}} \) & 1, 3 \\
\( S \) & 0.8, 1.0 \\
\( C \) & 0.8, 1.0 \\
\( \gamma \) & 0, 0.1 \\
\( \alpha \) & 0.5, 1 \\
\( \lambda \) & 0.5, 1 \\
\bottomrule
\end{tabular}
\end{table}

Table \ref{tab:param_grid} presents a parameter grid used in GridSearchCV, a technique for hyperparameter tuning in machine learning models. Hyperparameters are predefined settings that control the learning process of algorithms. The table lists various hyperparameters commonly used in the XGBoost regressor model, a popular gradient boosting framework \cite{chen2016xgboost}. Each hyperparameter is accompanied by its corresponding values, explored during the grid search process. For instance, \( N \) represents the number of estimators (trees) in the XGBoost model, with values of 300 and 400 being considered. Similarly, \( \eta \) denotes the learning rate, with potential values of 0.01, 0.1, and 0.2.

Other hyperparameters include \( D_{\text{max}} \) for maximum depth of trees, \( W_{\text{min}} \) for minimum child weight, \( S \) for subsampling ratio, \( C \) for column subsampling ratio, \( \gamma \) for minimum loss reduction required to make further splits, \( \alpha \) for L1 regularization term on weights, and \( \lambda \) for L2 regularization term on weights.

This parameter grid serves as a roadmap for systematically exploring various combinations of hyperparameters to identify the optimal configuration for the XGBoost model, thereby enhancing its predictive performance. The best combination of hyperparameters for the XGBoost model was selected based on the smallest RMSE, resulting in enhanced predictive performance. The chosen parameters are as follows:
\[
\Theta = \begin{pmatrix}
C & \gamma & \eta & D_{\text{max}} & W_{\text{min}} & N & \alpha & \lambda & S \\
1.0 & 0.1 & 0.1 & 4 & 3 & 400 & 0.5 & 1 & 0.8 \\
\end{pmatrix}
\]

Finally, the RMSE achieved with this parameter combination is the smallest observed during the hyperparameter tuning process.
\section{Results \& Interpretation} \label{sec:results_and_evaluation}

This section presents a comprehensive analysis of the results obtained from implementing various technical indicators, including Bollinger Bands, Average True Range, Commodity Channel Index, Williams \%R, and Chaikin Money Flow, on the Bitcoin price dataset. The performance and effectiveness of these indicators are evaluated through simulations, focusing on their ability to predict market trends and identify trading signals.

\subsection{Simulation Setup}
We conducted simulations on historical Bitcoin price data using a rolling window approach (15 minutes). Technical indicators (e.g., Bollinger Bands, Average True Range, Commodity Channel Index, Williams \%R, Chaikin Money Flow) were fed into the XGBoost Classifier, parameterized by the optimized vector $\Theta$ (described in Section \ref{sec:Mathematical_Model}), to generate buy/sell signals. Performance was evaluated using several metrics, such as accuracy, precision, and the ROC curve. Simulations were conducted on a Google Colab notebook using a T4 GPU.

\subsection{Results \& Analysis}
We present the experimental results that assess the performance of the proposed model, alongside a comparative analysis with the Logistic Regression model. The best parameters for Logistic Regression are shown in Appendix \ref{appendix_B}. 

\begin{table}[ht]
\centering
\caption{\textit{Comparison of Performance Metrics for XGBoost and Logistic Regression Models.}}
\label{tab:model_comparison}
\begin{tabular}{l c c}
\toprule
\textbf{Metric} & \textbf{XGBoost} & \textbf{Logistic Regression} \\
\midrule
Accuracy & 0.9240 & 0.9101 \\
Precision & 0.8917 & 0.8802 \\
Recall & 0.9490 & 0.9298 \\
F1 Score & 0.9195 & 0.9043 \\
ROC AUC & 0.9817 & 0.9760 \\
\bottomrule
\end{tabular}
\end{table}

Table \ref{tab:model_comparison} compares the key performance metrics of the XGBoost and Logistic Regression models. The metrics include Accuracy, Precision, Recall, F1 Score, and ROC AUC. Both models perform well, with XGBoost showing slightly better values across most metrics. XGBoost has a higher accuracy (0.9240 vs. 0.9101), recall (0.9490 vs. 0.9298), and ROC AUC (0.9817 vs. 0.9760), indicating better classification and discrimination capability. However, the Logistic Regression model still performs competitively, offering slightly lower but reasonable values for all metrics.



\begin{table}[!t]
    \centering
    \caption{Comparison of Confusion Matrices for XGBoost and Logistic Regression Models.}
    \label{tab:confusion_matrix_comparison}
    \begin{tabular}{@{}lcccc@{}}
        \toprule
        & \multicolumn{2}{c}{\textbf{XGBoost}} & \multicolumn{2}{c}{\textbf{Logistic Regression}} \\
        \cmidrule{2-5}
        \textbf{Actual} & \textbf{Predicted 0} & \textbf{Predicted 1} & \textbf{Predicted 0} & \textbf{Predicted 1} \\
        \midrule
        0 (Sell) & 3408 & 366 & 3372 & 402 \\
        1 (Buy)  & 162  & 3015 & 223  & 2954 \\
        \bottomrule
    \end{tabular}
\end{table}

Table \ref{tab:confusion_matrix_comparison} provides a comparison of the confusion matrices for the XGBoost and Logistic Regression models in predicting Bitcoin prices. The XGBoost model correctly identifies 3408 Sell operations and 3015 Buy operations, with 366 Sell operations misclassified as Buys and 162 Buy operations misclassified as Sells. In comparison, the Logistic Regression model correctly identifies 3372 Sell operations and 2954 Buy operations, with slightly higher misclassifications, as 402 Sell operations are misclassified as Buys and 223 Buy operations are misclassified as Sells.

Overall, the XGBoost model exhibits better performance in terms of correct classification, particularly with fewer misclassifications compared to Logistic Regression.
  
\begin{figure}[!t]
   \centering
   \includegraphics[width=3.5in]{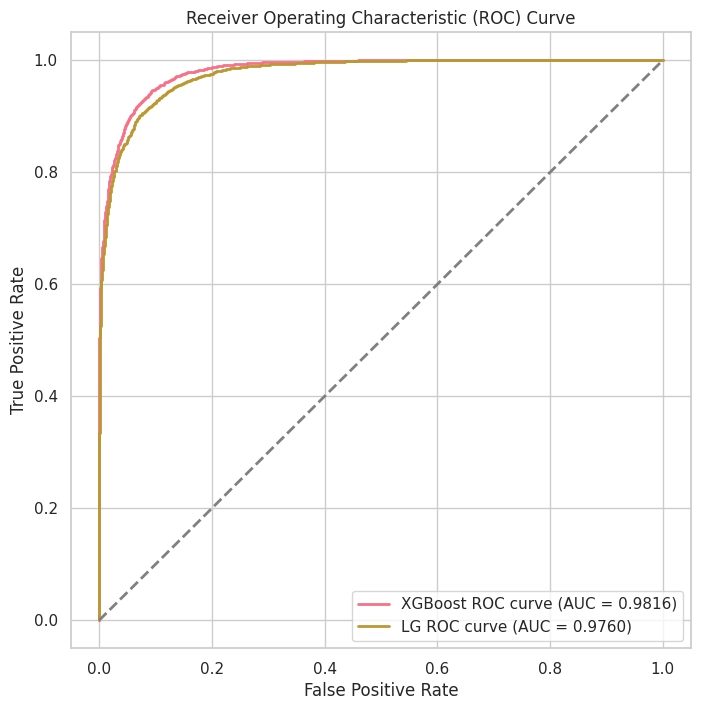}
   \caption{Receiver Operating Characteristic (ROC) curve comparison for XGBoost and Logistic Regression.}
   \label{fig_roc_comparison}
\end{figure}

Figure \ref{fig_roc_comparison} illustrates the Receiver Operating Characteristic (ROC) curves for both the XGBoost and Logistic Regression models. The ROC curve visually represents the trade-off between the true positive rate (recall/sensitivity) and the false positive rate (1-specificity) at various threshold levels. Notably, the XGBoost curve is positioned closer to the ideal top-left corner compared to Logistic Regression, indicating superior discriminative ability between the positive and negative classes. The area under the curve (AUC) for XGBoost is higher, signifying better overall performance, with a more effective balance of minimizing false positives while maximizing true positives.

\begin{figure*}[!t]
   \centering
   \includegraphics[width=\textwidth]{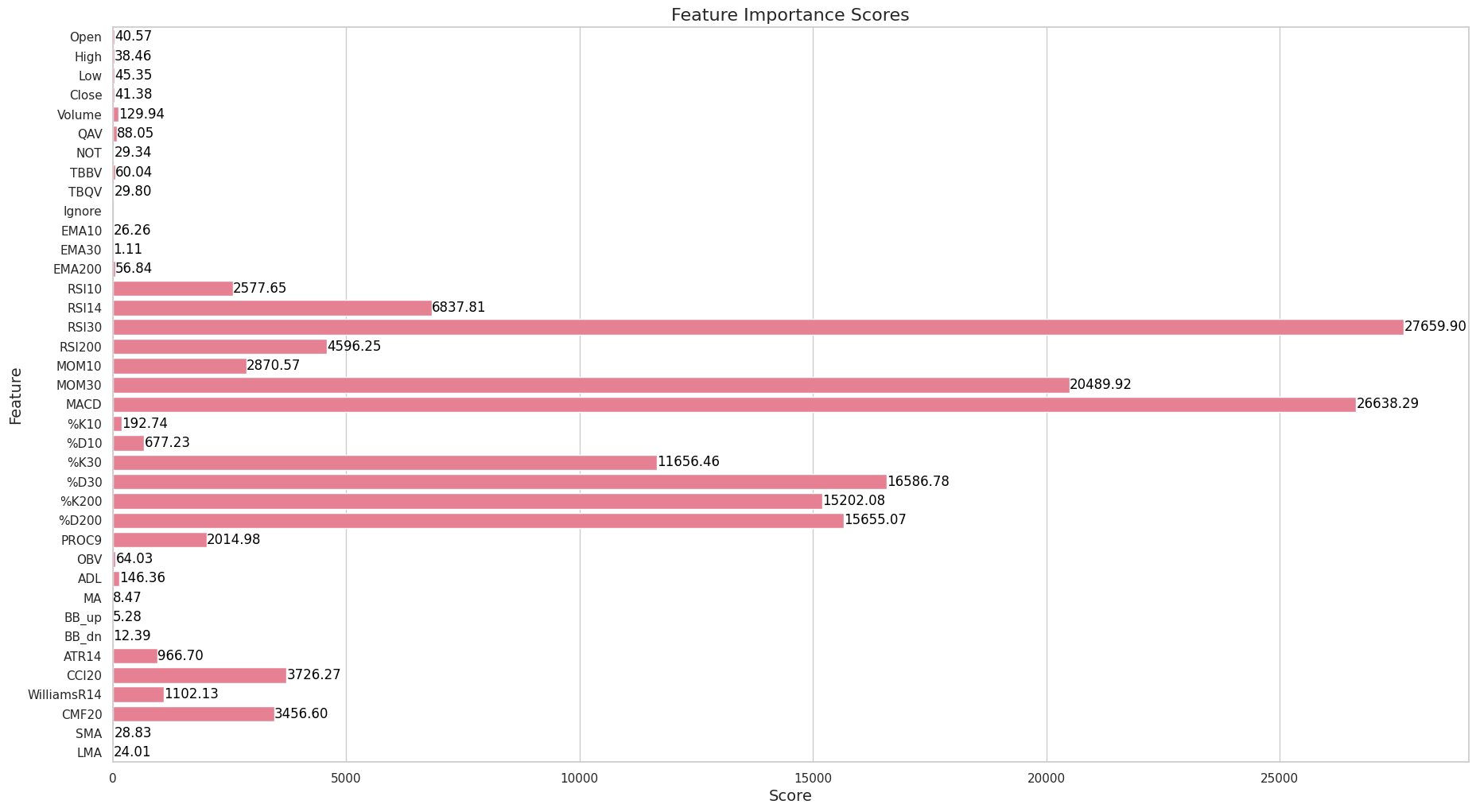}
   \caption{Importance scores of various technical indicators and historical data determined by the \(\chi^2\) statistical test.}
   \label{fig:feature_importance}
\end{figure*}

Figure \ref{fig:feature_importance} shows the importance scores of various technical indicators and historical data determined by the \(\chi^2\) statistical test. The figure illustrates the relative significance of these features in predicting Bitcoin prices. The top 8 features, based on their importance scores, are RSI30, MACD, MOM30, \%D30, \%D200, \%K200, \%K30, and RSI14, highlighting their critical role in the feature selection process for the predictive model.

\begin{figure}[!t]
   \centering
   \includegraphics[width=3.5in]{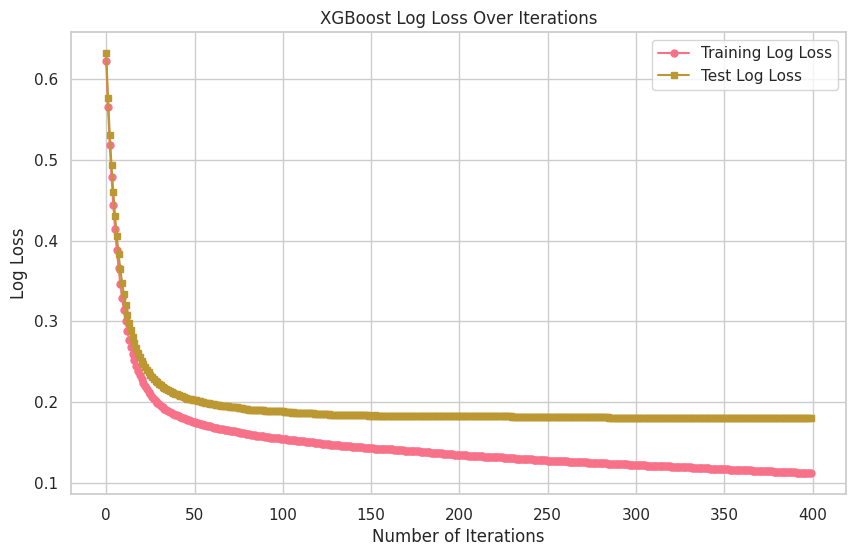}
   \caption{Log loss over iterations for training and test sets. The figure demonstrates the behavior of the log loss function for the XGBoost model as the number of iterations increases.}
   \label{fig_loss}
\end{figure}

Figure \ref{fig_loss} illustrates the log loss function for both the training and test sets over the number of iterations. The log loss, a standard metric for evaluating the performance of classification models, measures the uncertainty of predictions. As shown in the figure, the log loss decreases steadily with the increase in the number of iterations, indicating that the model is learning effectively.

Specifically, the training and test log loss values decrease at a similar rate, and they remain close to each other throughout the iterations. This close alignment between the training and test log loss curves is a positive indicator, suggesting that the model is generalizing well to unseen data and is not overfitting.

The decreasing trend of the log loss function for both sets confirms that the model's predictions are becoming more accurate with more iterations. Additionally, the absence of a significant gap between the training and test log loss values suggests that the regularization techniques employed effectively prevent overfitting. This balance is crucial for ensuring the model's robustness and reliability in practical applications.

\begin{figure}[!t]
   \centering
   \includegraphics[width=3.5in]{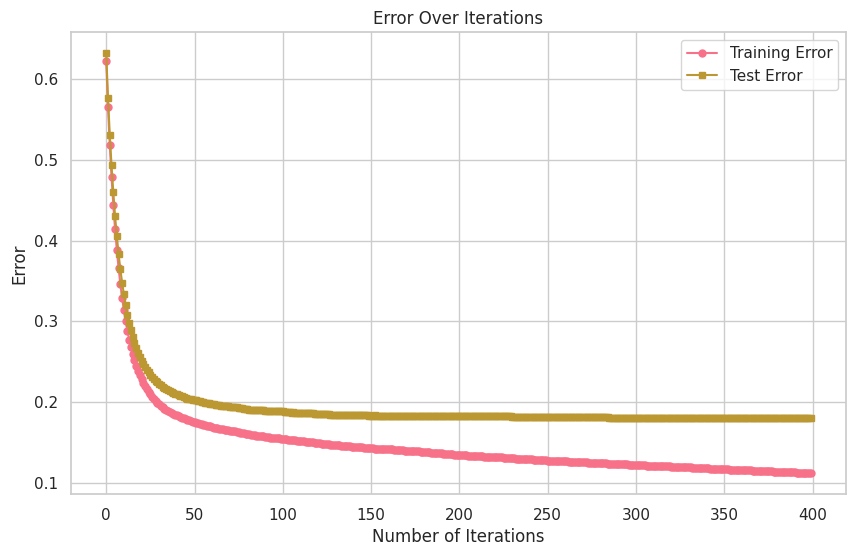}
   \caption{Error over iterations for training and test sets. The figure shows the behavior of the error function for the XGBoost model as the number of iterations increases.}
   \label{fig_error}
\end{figure}

Figure \ref{fig_error} illustrates the error function for both the training and test sets over the number of iterations for the XGBoost model. The error function measures the proportion of incorrect predictions made by the model. Mathematically, the error rate at iteration \( t \) can be expressed as:

\[
\text{Error}_t = \frac{1}{m} \sum_{i=1}^{m} \mathbf{1}(\hat{y}_i^{(t)} \neq y_i),
\]

where \( \hat{y}_i^{(t)} \) is the predicted label for the \( i \)-th sample at iteration \( t \), \( y_i \) is the true label, and \( \mathbf{1}(\cdot) \) is the indicator function that equals 1 when its argument is true and 0 otherwise.

As depicted in the figure, the error decreases consistently as the number of iterations increases, indicating that the model's performance improves with more iterations. The training and test error values decline at a similar pace, and they remain close throughout the iterations.

The convergence of the training and test error curves suggests that the model effectively learns from the training data and generalizes well to the test data. The small gap between the training and test error values indicates that the model is not overfitting, which is corroborated by regularization techniques.

The error function's downward trend signifies the model's increasing accuracy over time, with fewer misclassifications occurring as the number of iterations rises. This close alignment between the training and test error curves further supports the robustness and reliability of the XGBoost model in making accurate predictions.


\section{Discussion \& Comparison}
\label{sec:discussion_and_comparison}

\begin{table*}[ht]
\centering
\caption{\textit{Comparison of current work with existing studies based on various factors including the focus of the paper, the methodologies employed, different metrics utilized, and the resultant outcomes.}}
\label{tab:comparison}
\begin{tabular}{p{3cm} p{2cm} p{4cm} p{2cm} p{3cm}}
\toprule
\textbf{Study} & \textbf{Focus} & \textbf{Methods} & \textbf{Metrics} & \textbf{Results} \\
\midrule
Shynkevich et al. (2017) \cite{shynkevich2017forecasting} & Stock prices & Technical indicators, ML algorithms & Accuracy, Sharpe ratio & Optimal window $\approx$ horizon \\
Liu et al. (2021) \cite{liu2021forecasting} & Bitcoin prices & SDAE, BPNN, SVR & MAPE, RMSE, DA & SDAE: MAPE 4.5\%, RMSE 0.012 \\
Jaquart et al. (2022) \cite{jaquart2022machine} & Crypto trading & LSTM, GRU & Accuracy, Sharpe ratio & Accuracy: 57.5\%-59.5\%, SR: 3.23 (LSTM) \\
Saad et al. (2019) \cite{saad2019toward} & Crypto dynamics & Economic theories, ML methods & Accuracy & Up to 99\% accuracy \\
Akyildirim et al. (2021) \cite{akyildirim2021prediction} & Crypto prices & SVM, Logistic Regression, ANN, RF & Accuracy & Accuracy: 55\%-65\% \\
Roy et al. (2023) \cite{roy2023forecasting} & Bitcoin prices & LSTM & MAE, RMSE, $R^{2}$ & MAE: 253.30, RMSE: 409.41, $R^{2}$: 0.9987 \\
Hafid et al. (2023) \cite{hafid2023bitcoin} & Bitcoin trends & RF, technical indicators & Accuracy & 86\% accuracy \\
Passalis et al. (2018) \cite{passalis2018temporal} & Stock prices & Temporal BoF model & Precision, Recall, F1 score, Cohen's $\kappa$ & Precision: 46.01, Recall: 56.21, F1 score: 45.46, $\kappa$: 0.2222 \\
Lin et al. (2020) \cite{lin2020forecasting} & Stock prices & RNN & RMSE, MAE, Accuracy & RMSE: 12.933, MAE: 10.44, Accuracy: 59.4\% \\
\textbf{Current Work} & Bitcoin trading & XGBoost, chi-squared test, technical indicators & Accuracy, feature importance & Key features: RSI30, MACD, MOM30, Accuracy: 92.4\% \\
\bottomrule
\end{tabular}
\end{table*}

Table \ref{tab:comparison} compares the current work with several significant existing stock and cryptocurrency price prediction studies. This table outlines each study's focus, methods, metrics, and key results, allowing for a clear comparison of different approaches and their outcomes. Each study is categorized based on its primary focus, ranging from stock prices to cryptocurrency dynamics, and employs various methods such as machine learning algorithms, economic theories, and technical indicators. Key performance metrics include Acc., SR, MAPE, RMSE, MAE, and $R^{2}$, providing a thorough evaluation framework across different prediction models. The results highlight the effectiveness of each approach in forecasting price movements and offer insights into the progress made in predictive modeling within the financial and cryptocurrency sectors.

The analysis highlights the strengths and limitations of each technical indicator. For instance, while Bollinger Bands effectively captured price reversals, it occasionally generated false signals in highly volatile markets. Similarly, the ATR helped adjust trading strategies based on market volatility but required careful tuning of the parameter \(\tau\). The CCI and Williams \%R indicators were sensitive to sudden price changes, making them suitable for short-term trading strategies. CMF provided a unique perspective on market sentiment by incorporating volume data, enhancing the overall analysis.

\section{Conclusion} \label{sec:conclusion}

With an emphasis on Bitcoin cryptocurrency, we have introduced a classification-based machine learning model in this work to forecast the direction of cryptocurrency markets. By applying crucial technical indicators like the Moving Average Convergence Divergence and Relative Strength Index, we were able to train a machine learning model to produce remarkably accurate buy and sell recommendations.

Our thorough empirical investigation shows that the suggested methodology works well. The robustness and dependability of the model are demonstrated by the classification report's accuracy of 92.40\%, precision of 89.17\%, recall of 94.90\%, F1 score of 91.95\%, and ROC AUC of 98.17\%. The confusion matrix and the ROC curve further support the model's remarkable ability to differentiate between positive and negative classes.

The importance of technical indicators in predicting market trends and identifying trading signals is highlighted by analyzing several of them, such as Bollinger Bands, Average True Range, Commodity Channel Index, Williams \%R, and Chaikin Money Flow. The relevance of indicators like RSI30, MACD, MOM30, \%D30, \%D200, \%K200, \%K30, RSI14, RSI200, and CCI20 in the feature selection process is highlighted by the critical scores of the \(\chi^2\) statistical test.

Furthermore, with a steady decline in both log loss and error rates, our examination of the log loss and error functions over iterations shows that the XGBoost model learns from the training data efficiently and generalizes well to new data. The slightest difference between test and training metrics indicates the model's resilience and capacity to prevent overfitting.

Our machine learning model shows significant potential in aiding cryptocurrency traders and investors in making informed decisions in the volatile and dynamic market. If more market indicators are added and various machine learning algorithms are tested, future studies should investigate how to improve prediction accuracy and robustness even more.

\section*{Data Availability}
The data that support the findings of this study are available from the corresponding author upon reasonable request.

\appendices

\section{Time Series Data Splitting for Training and Testing}
\label{appendix_A}

This appendix presents a mathematical expression for splitting our time series dataset into training and testing sets. The objective is to allocate a specified percentage for the testing set while maintaining the order of observations.

Let \( p \) be the desired percentage of the dataset for the testing set.
\subsection{Training Set (First Part)}
The training set includes the initial part of the time series, starting from the first observation and continuing up to a specified index \( t_{\text{train\_end}} \). The training set indices are given by:
\[ T_{\text{train}} = \{1, 2, 3, \ldots, t_{\text{train\_end}}\} \]

\subsection{Testing Set (Remaining Part with Percentage Adjustment)}
\label{subsec:test_set}
The testing set includes the remaining part of the time series, starting from the index \( t_{\text{test\_start}} + 1 \) and continuing up to the last observation. The testing set indices are given by:
\[ T_{\text{test}} = \{t_{\text{test\_start}} + 1, t_{\text{test\_start}} + 2, \ldots, m\} \]
where \( t_{\text{test\_start}} = \lfloor (1 - p) \cdot m \rfloor \).

This approach ensures that the testing set is adjusted to the desired percentage of the dataset. The parameter \( p \) can be adjusted based on specific requirements.
\section{Hyperparameter Tuning for Logistic Regression}
\label{appendix_B}

We employed GridSearchCV for hyperparameter optimization of our logistic regression model. The parameter space explored is presented in Table \ref{tab:hyperparam_grid}.

\begin{table}[ht]
\centering
\begin{tabular}{ll}
\hline
\textbf{Hyperparameter} & \textbf{Values} \\
\hline
Penalty & $\ell_1$, $\ell_2$, elasticnet, none \\
C & 0.01, 0.1, 1, 10, 100 \\
Solver & liblinear, saga \\
max\_iter & 100, 200, 300 \\
\hline
\end{tabular}
\caption{Hyperparameter grid for logistic regression}
\label{tab:hyperparam_grid}
\end{table}

This systematic search aims to identify the optimal configuration, enhancing the model's predictive performance and generalization capabilities  (i.e., preventing overfitting). After hyperparameter tuning, the best parameters found for the logistic regression model are \texttt{params\_log\_reg = \{`C': 0.1, `max\_iter': 100, `penalty': `l1', `solver': `saga'\}}.
%
%
\bibliographystyle{IEEEtran}
\bibliography{bibliography}
\end{document}